\documentclass[runningheads]{llncs}
\usepackage{graphicx}
\usepackage{times}
\usepackage{latexsym}

\usepackage[T1]{fontenc}

\usepackage[utf8]{inputenc}

\usepackage{microtype}

\usepackage{helvet} 
\usepackage{courier}  
\usepackage{url}  
\usepackage{graphicx} 
\usepackage{wrapfig}
\usepackage[utf8]{inputenc}
\usepackage{subfigure}
\usepackage{booktabs}
\usepackage{amsmath}
\usepackage{amsfonts}
\usepackage{multirow}
\usepackage{color}
\usepackage[lined,boxed,commentsnumbered, ruled,linesnumbered]{algorithm2e}
\usepackage{enumitem}
\usepackage{microtype}
\usepackage{tabularx}
\usepackage{arydshln}
\usepackage{chngpage}
\usepackage{array}
\usepackage{hyperref}

\usepackage{microtype}
\usepackage{makecell}
\usepackage{tabularx}
\usepackage{booktabs}

\usepackage{marvosym}

\begin{document}
\title{Extractor-enhanced Aspect Controllable Summarization for E-commerce Products}
\title{Variational Aspect-Oriented Attention for \\ Diverse Product Summarization}
\title{Aspect-aware Copywriting: Describing Product for E-Commerce}
\title{CUSTOM: Aspe\underline{c}t-Oriented Prod\underline{u}ct \underline{S}ummariza\underline{t}ion for E-C\underline{om}merce}

%
\titlerunning{CUSTOM}
%

\author{Jiahui Liang\thanks{Corresponding author} \and
Junwei Bao \and
Yifan Wang \and
Youzheng Wu \and
Xiaodong He \and
Bowen Zhou}

\authorrunning{J. Liang et al.}
%
\institute{JD AI Research, Beijing, China \\
\email{\{liangjiahui14,baojunwei,wangyifan15\}@jd.com}\\
\email{\{wuyouzhen1,xiaodong.he,bowen.zhou\}@jd.com}\\
}
%

\maketitle              
\begin{abstract}
Product summarization aims to automatically generate product descriptions, which is of great commercial potential.
Considering the customer preferences on different product aspects, it would benefit from generating aspect-oriented customized summaries.
However, conventional systems typically focus on providing general product summaries, which may miss the opportunity to match products with customer interests.
To address the problem, we propose CUSTOM, aspe\underline{c}t-oriented prod\underline{u}ct \underline{s}ummariza\underline{t}ion for e-c\underline{om}merce, which generates diverse and controllable summaries towards different product aspects.
To support the study of CUSTOM and further this line of research, we construct two Chinese datasets, i.e., {S}{\small MART}{P}{\small HONE} and {C}{\small OMPUTER}, including 76,279 / 49,280 short summaries for 12,118 / 11,497 real-world commercial products, respectively.
Furthermore, we introduce EXT, an \underline{ext}raction-enhanced generation framework for CUSTOM, where two famous sequence-to-sequence models are implemented in this paper.
We conduct extensive experiments on the two proposed datasets for CUSTOM and show results of two famous baseline models and EXT, which indicates that EXT can generate diverse, high-quality, and consistent summaries.\footnote{\url{https://github.com/JD-AI-Research-NLP/CUSTOM}}

\end{abstract}

\section{Introduction}
\noindent 
Product summarization aims to automatically generate product descriptions for e-commerce.
It is of great commercial potential to write customer-interested product summaries.
In recent years, a variety of researches focus on product summarization and have proposed practical approaches~\cite{documentSummary,aspectAware,longDiverseGeneration,automaticGeneration}.
These models take product information as input and output a general summary describing a product. 
However, the general summaries usually face two problems: (1) they are singular and lack diversity, which may miss the opportunity to match products with customer interests; (2) it is not able to control the models to describe what aspects of the products in the general summaries, which hurts for personalized recommendations.
Considering that different customers have preferences on different aspects of products on e-commerce platforms, e.g., the customers may care more about the ``APPEARANCE" than the ``PERFORMANCE" aspect of a smartphone, so generating aspect-oriented customized summaries will be beneficial to personalized recommendations.

\begin{figure*}[t]
	\centering
	\includegraphics[width=4.4in]{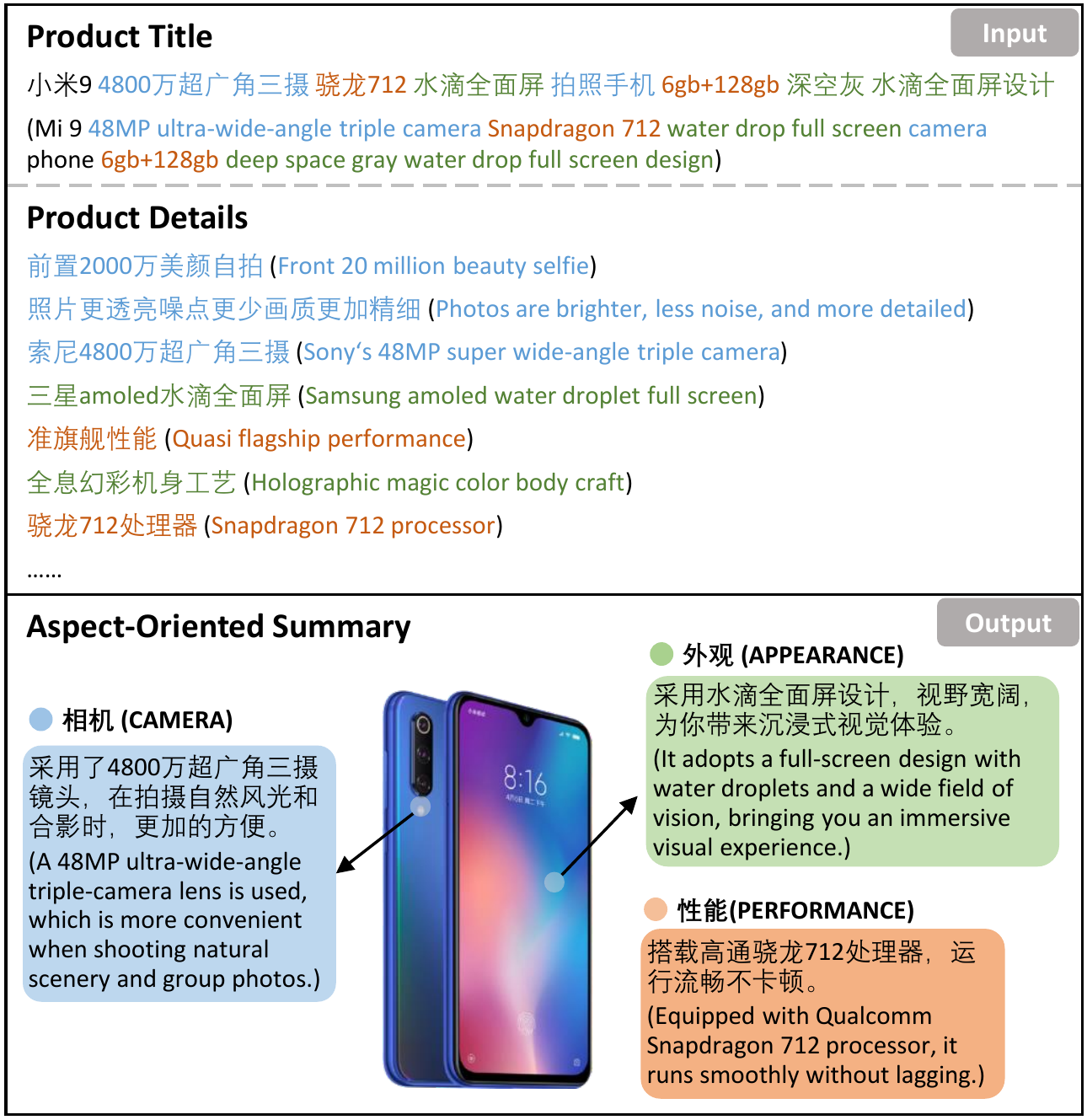}
	\caption{\label{task} An example of CUSTOM: aspe\underline{c}t-oriented prod\underline{u}ct \underline{s}ummariza\underline{t}ion for e-c\underline{om}merce.}
\end{figure*}

Motivated by the above issues and observation, in this paper, we propose CUSTOM, aspe\underline{c}t-oriented prod\underline{u}ct \underline{s}ummariza\underline{t}ion for e-c\underline{om}merce, which can provide diverse and controllable summaries towards different product aspects.
In detail, given a product with substantial product information and a set of corresponding aspects for this category, e.g., smartphone and computer, the task of CUSTOM is to generate a set of summaries, each of which only describes a specified aspect of the product.
Figure~\ref{task} shows an example of a smartphone where the product information includes a product title and substantial product details in a natural language recognized from abundant product images.
Different colors in the product information represent different aspects of the product.
When a certain aspect, i.e., CAMERA, APPEARANCE, or PERFORMANCE, is specified, the corresponding aspect-oriented summary is generated based on the product information. 
For example, the ``green" summary describes the appearance of the smartphone.
To the best of our knowledge, our task is related to the conditional generation~\cite{kobe,controllableSummarization,controlledGeneration,headlineGeneration}.
Specifically, KOBE~\cite{kobe} is the most similar task to CUSTOM.
KOBE focuses on writing expansion from short input, i.e., product title, where the generated text is likely to disrespect the truth of products, while CUSTOM concentrates on generating summary from long input, i.e., substantial product information, where the generated summary is consistent with the truth of products.

To support the study of CUSTOM and further this line of research, we construct two Chinese datasets, i.e., {S}{\small MART}{P}{\small HONE} and {C}{\small OMPUTER}, including 76,279 / 49,280 short summaries for 12,118 / 11,497  real-world commercial products, respectively.
Furthermore, inspired by the content selection methods proposed in \cite{mixtureSelection,bottomup,unifiedModel,selectivesummary}, we introduce EXT, an \underline{ext}raction-enhanced generation framework for CUSTOM, which equips the model with the ability to select aspect-related sentences from product information to enhance the correlation between the generated summary and the aspect.
We conduct extensive experiments on the two proposed datasets and show results of Pointer-Generator, UniLM, and EXT, which indicate that EXT can generate more diverse, high-quality, and consistent summaries.
Our contributions are as follows:
\begin{itemize}
	\setlength{\itemsep}{0pt}
	\setlength{\parsep}{0pt}
	\setlength{\parskip}{0pt}
\item We propose CUSTOM, aspe\underline{c}t-oriented prod\underline{u}ct \underline{s}ummariza\underline{t}ion for e-c\underline{om}merce, to generate diverse and controllable summaries towards different product aspects.
\item We construct two real-world Chinese commercial datasets, i.e., {S}{\small MART}{P}{\small HONE} and {C}{\small OMPUTER}, to support the study of CUSTOM and further this line of research.
\item We introduce EXT, an \underline{ext}raction-enhanced generation framework. Experiment results on {S}{\small MART}{P}{\small HONE} and {C}{\small OMPUTER} show the effectiveness of the EXT.
\end{itemize}

\section{Methodology}
\subsection{CUSTOM: Aspe\underline{c}t-Oriented Prod\underline{u}ct \underline{S}ummariza\underline{t}ion for E-C\underline{om}merce\label{custom}}
In this section, we formulate the CUSTOM task as follows.
With respect to Figure~\ref{task}, the input of the model includes product information ${S}$ and an aspect category ${a}$, where ${S}$=$\{s_i\}_{i=1}^{N}$ is a series of sentences by concatenating the title and product details, ${N}$ represents the maximum number of sentences. The output is a short summary ${y}$ that only describes the specified aspect ${a}$ of the product.

\subsection{{S}{\small MART}{P}{\small HONE} and {C}{\small OMPUTER}}
\begin{figure*}[h]
	\centering
	\includegraphics[width=4.8in]{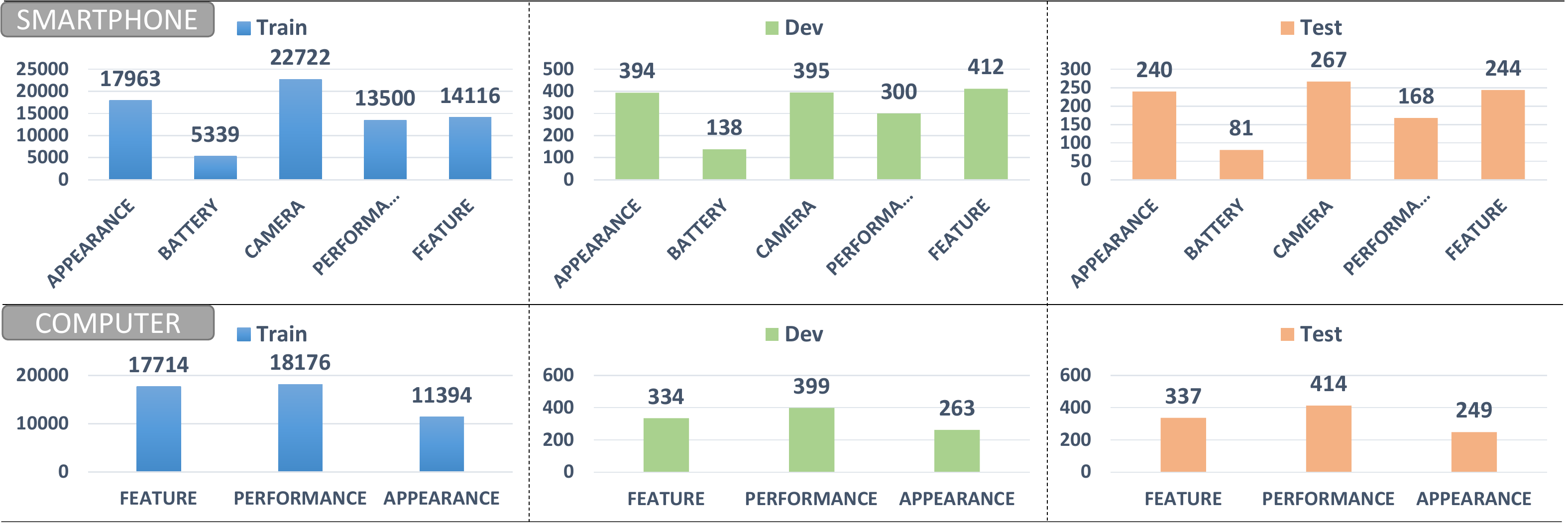}
	\caption{\label{aspectDist} Aspect Distributions for {S}{\small MART}{P}{\small HONE}  and {C}{\small OMPUTER}.}
\end{figure*}

To support the study of CUSTOM and further this line of research, we construct two Chinese datasets, i.e., {S}{\small MART}{P}{\small HONE} and {C}{\small OMPUTER}, including 76,279 / 49,280 short summaries for 12,118 / 11,497 real-world commercial products, respectively.
We split the two datasets into training, development, and test for experiments.
The number of instances and products of train, development, and test sets are shown in Table~\ref{dataset}.
As described in Section~\ref{custom}, each dataset is consist of a set of $\langle$\textit{product information}, \textit{aspect}, \textit{aspect summary}$\rangle$ tuples.
In the {S}{\small MART}{P}{\small HONE} training set, the average number of different aspect summaries for each product is about 6.3, while it is 1.8 and 2.1 in development and test sets respectively. The reason is that we keep the pair of $\langle$\textit{product information}, \textit{aspect}$\rangle$ unique in the development and test sets in order to ensure that an input $\langle$\textit{product information}, \textit{aspect}$\rangle$ pair corresponds to an unique output summary. The same is true for {C}{\small OMPUTER} dataset.
For {S}{\small MART}{P}{\small HONE} dataset, there are five kinds of aspects, i.e., \textit{APPEARANCE}, \textit{BATERY}, \textit{CAMERA}, \textit{PERFORMANCE}, and \textit{FEATURE}.
For {C}{\small OMPUTER} dataset, there are three kinds of aspects, i.e., \textit{FEATURE}, \textit{PERFORMANCE}, \textit{APPEARANCE}.
Figure~\ref{aspectDist} shows the aspect distributions on training, development, and test set for {S}{\small MART}{P}{\small HONE} and {C}{\small OMPUTER}.

\begin{table}[h]
\centering
\small
\caption{Statistics of {S}{\small MART}{P}{\small HONE} and {C}{\small OMPUTER}. \#sum: the number of aspect summaries. \#prod: the number of products. \label {dataset}}
\setlength\tabcolsep{4pt}
\begin{tabular}{lcccccccc}
  \hline 
  {\bf Category} & \multicolumn{2}{c} {\bf Overall} & \multicolumn{2}{c} {\bf Train} & \multicolumn{2}{c} {\bf Dev} & \multicolumn{2}{c} {\bf Test} \\
  & \#sum & \#prod & \#sum & \#prod & \#sum & \#prod & \#sum & \#prod\\
  \hline 
  {S}{\small MART}{P}{\small HONE} & 76,279 & 12,118 & 73,640 & 10,738 & 1,639 & 896 & 1,000 & 484 \\
  \hline 
  {C}{\small OMPUTER} & 49,280 & 11,497 & 47,284 & 10,268 & 996 & 615 & 1,000 & 614 \\
  \hline
\end{tabular}
\end{table}

The datasets are collected from a Chinese e-commerce platform and we have summaries written by professional writers that describe multiple aspects of the product.
Since there are no $\langle$\textit{product information}, \textit{aspect}, \textit{aspect summary}$\rangle$ tuples, we use a heuristic method to construct the datasets. 
First, we split the origin summary by periods, and only keep fragments with character lengths between 15 and 55. 
The shorter fragments may contain insufficient information while the longer fragments could contain more than one aspect description. 
We then use BERT~\cite{bert} to obtain continuous vectors of each fragment and used them as input to the K-Means~\cite{kmeans} clustering algorithm. 
The resulting clusters represent the description set of different aspects of the product, that is, the aspect summary set.
The cluster label of each instance is the aspect category. 
We experiment with different numbers of clusters, and manually select 100 from each clustering result for evaluation, and select the result with the highest degree of separation of the clustering results as the final number of clusters. 
We have also tried some methods to automatically detect clustering results, such as the silhouette coefficient, but the results are not ideal. 
Finally, we find that the best number of clusters is 5 and 3 for the {S}{\small MART}{P}{\small HONE} and {C}{\small OMPUTER}, respectively.

\begin{figure*}[tb]
	\centering
	\includegraphics[width=4.6in]{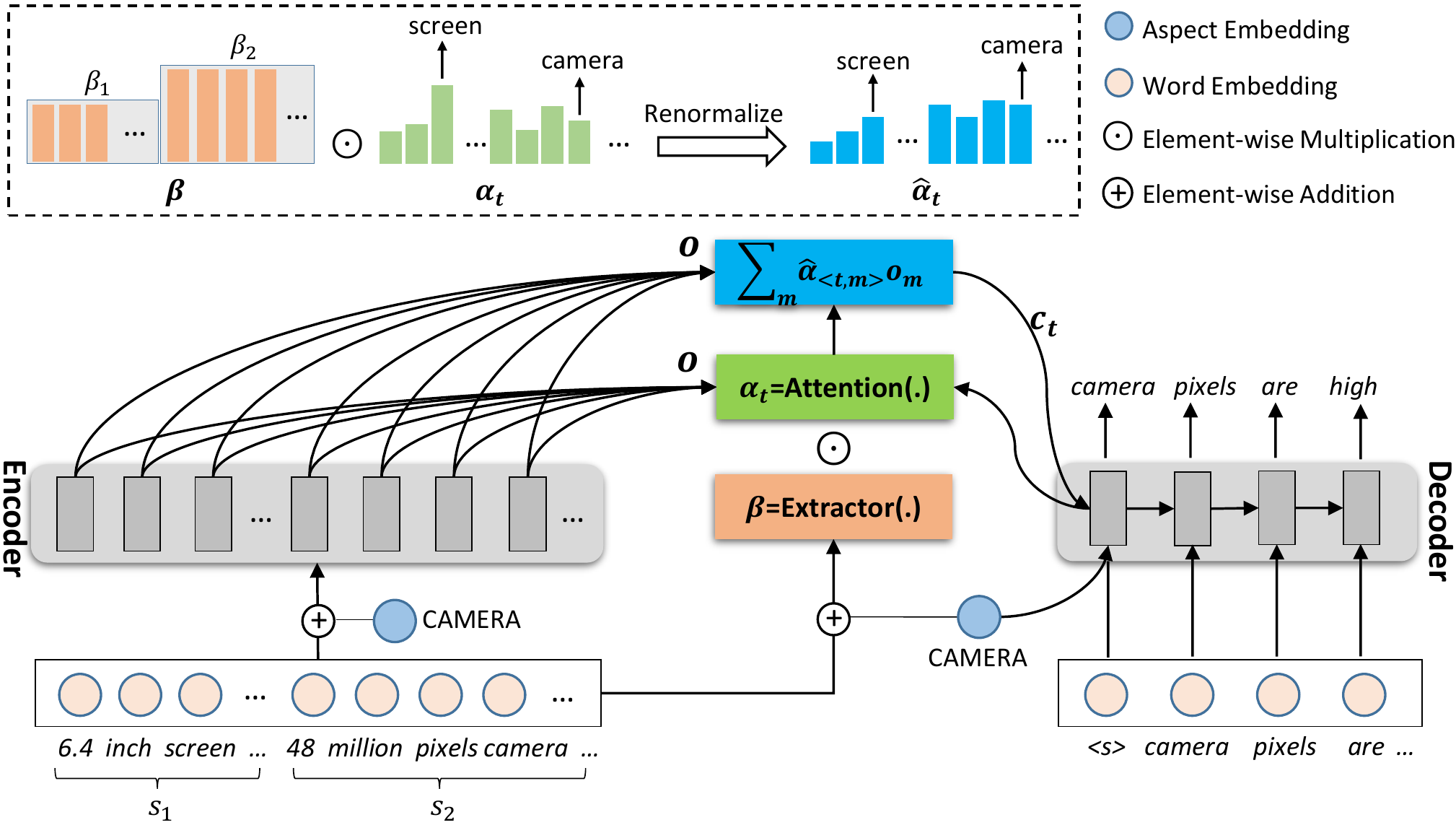}
	\caption{\label{model} The proposed \underline{ext}raction-enhanced generation (EXT) framework.}
\end{figure*}
\subsection{EXT: \underline{Ext}raction-Enhanced Generation Framework \label{extSection}}
We introduce EXT, an extraction-enhanced generation framework for CUSTOM.
As shown in Figure \ref{model}, the framework is consists of two parts: extractor and generator.

The extractor is responsible for scoring each sentence in the product information and outputs the sentence-level attention:
\begin{equation}
\mathbf{\beta_i}=\mathtt{Ext}(\mathbf{s}_{i},\mathbf{a})
\end{equation}
where ${{s}_{i}}$ is the ${i^{th}}$ sentence and ${a}$ is the input aspect category, the greater the ${\beta_i}$, the more relevant this sentence is to the current aspect, and vice versa. The binary cross-entropy loss function is used to train the extractor:
\begin{equation}
L_{ext}=-\frac{1}{N} \sum_{i=1}^{N}\left(g_{i} \log \beta_{i}+\left(1-g_{i}\right) \log \left(1-\beta_{i}\right)\right)
\end{equation}
where ${g_{i}}$ is the aspect label for the ${i^{th}}$ sentence. ${g_{i}}$ is 1 means that the description of the current sentence is consistent with the current aspect.
We use a heuristic rule to obtain the aspect labels.
Specifically, we calculate the overlap rate $r=\frac{\#overlap\_length}{\#sentence\_length}$ between each sentence in the product information and the aspect summary, where $\#overlap\_length$ is calculated by longest common subsequence (LCS)~\cite{LCS}. The aspect label is 1 if the overlap rate is above a certain threshold, which is 0.35 in our setting.

The generator is basically a sequence-to-sequence structure with attention mechanism \cite{bahdanau2014neural}. 
Inspired by the study~\cite{unifiedModel}, at each decoding step ${t}$, we combine the sentence-level score ${\beta_i}$ generated by the extractor and word-level attention ${{\alpha}_{m}^{t}}$ at time step ${t}$ as follows:
\begin{equation}
    \hat{\alpha}_{m}^{t}=\frac{\alpha_{m}^{t} \times \beta_{i(m)}}{\sum_{m} \alpha_{m}^{t} \times \beta_{i(m)}} \label{combine_attention}
\end{equation}
where ${m}$ is the word index.
The updated word-level attention weight $\hat{\alpha}_{m}^{t}$ is then used to compute context vector ${c}_{t}$ as follows:
\begin{equation}
    c_{t}=\sum_{m=1}^{|w|} \hat{\alpha}_{m}^{t} o_{m}
\end{equation}
where ${o_{m}}$ stands for the ${m^{th}}$ word representation.

We reuse the aspect information and fuse it to the generator to calculate the final word distribution ${P(w)}=\mathtt{Gen}(y_{t-1}, \mathbf{c}_{t}, \mathbf{a})$ at step ${t}$.
We train the generator with the average negative log-likelihood loss: 
\begin{equation}
L_{gen}=-\frac{1}{T} \sum_{t=1}^{T} \log P\left(y_{t}\right)
\end{equation}
where ${y_{t}}$ is the ground-truth target word.
The extractor and generator are trained end-to-end, the final loss is as below:
\begin{equation}
L=L_{ext}+L_{gen}
\end{equation}

We apply this framework to two typical summarization models, one is an RNN-based end-to-end model called Pointer-Generator~\cite{see-etal-2017-get}, and the other is a Transformer-based pretrained model called UniLM~\cite{dong2019unified}. We noticed that UniLM, T5~\cite{t5}, and BART~\cite{bart} are all SOTAs of pretrained generative models. In this paper, we choose UniLM since it is representative and only UniLM has released the Chinese version pretrained parameters.
\subsubsection{EXT on Pointer-Generator}
First, we concatenate each sentence in ${S}$ with a period to form a sequence of words $\{w_m\}_{m=1}^{|w|}$.
Then each word ${w_{m}}$ and the aspect category ${a}$ are embedded to obtain continuous vector ${e_{{w}_{m}}}$ and ${e_{a}}$ respectively. 
The word embedding and aspect embedding are added by column to obtain the fused aspect-aware word embedding ${e_{{f}_{m}}}$.
The fused embedding is used as initial input for both extractor and generator.

For extractor, we use a hierarchical bidirectional GRU as in \cite{hierarchicalGru} to extract sentence representations $\mathbf{H}$. Then we apply a matrix dot operation and uses the sigmoid function to predict the sentence-level attention ${\beta_i}$:
\begin{equation}
\mathbf{\beta_i}=\operatorname{Sigmoid}(\mathbf{{e}_{a}}^\mathrm{T} \mathbf{{h}_{i}})
\end{equation}
where ${\mathbf{{h}_{i}}}$ stands for ${i^{th}}$ sentence vector.

The generator is basically the pointer-generator network and we combine the sentence-level attention and word-level attention as described in equation \eqref{combine_attention}. 
Besides being used to compute context vector ${c}_{t}$, the updated word attentions are also used as the new copy distribution.
The aspect embedding is also added to the previous word embedding by column at each decoding step.
\subsubsection{EXT on UniLM}
We use the same fused embedding ${e_{{f}_{m}}}$ as input, except ${e_{{w}_{m}}}$ is the summation of 3 kinds of embeddings including word embedding, position embedding, and segment embedding.
The input embeddings are encoded into contextual representations $\mathbf{O}=[{\mathbf{o_{1}}, \dots, \mathbf{o_{|w|}}}]$ using Transformer.
Since the sentences are separated by periods, we simply take out the hidden vector of each period to represent the sentence representation $\mathbf{{H}}$:
\begin{equation}
\mathbf{H}=\mathbf{O}[{I}_{1}, \dots, {I}_{M}]
\end{equation}
where $\mathbf{{I}_{{i}}}$ denotes the index of period of sentence $\mathbf{{s}_{i}}$. Then we apply a two-layer feedforward network to compute the sentence score:
\begin{equation}
\mathbf{\beta_i}=\operatorname{Sigmoid}(\mathbf{FFN}(\mathbf{{h}_{i}}))
\end{equation}
The sentence scores are used to update word-level attention as in equation \eqref{combine_attention}.
The generator shares the parameters of the transformer encoder with the extractor.
Finally, we apply the same fine-tuning method as in UniLM~\cite{dong2019unified}.
\section{Experiment}

\subsection{Comparison Methods}
In this section, we note the baselines and our proposed methods.
\textbf{PGen w/o Aspect} and \textbf{UniLM w/o Aspect} denotes the original Pointer-Generator Network and UniLM model without aspect information as input.
\textbf{PGen w/ Aspect} and \textbf{UniLM w/ Aspect} means we fuse aspect information as describe in Section~\ref{extSection} but without incorporating \textbf{EXT}.
\textbf{EXT-PGen} denotes our proposed PGen-based model.
\textbf{EXT-UniLM} represents our proposed UniLM-based model.

\subsection{Implementation Details}
For PGNet-based models, the model parameters remain the same as work \cite{unifiedModel}. 
For UniLM-based models, we use the base version and load the pre-trained parameters published in work\footnote{\url{https://github.com/YunwenTechnology/Unilm}} for initializing. 
The maximum numbers of the characters in the input and target are 400 and 70, respectively.
During training, the mini-batch size is set to 20.
We choose the model with the smallest perplexity on the development dataset for evaluation.
During inference, the beam size is set to 5, and the maximum decoding length is set to 80 for all models.

\subsection{Diversity Evaluation for CUSTOM}
We argue that CUSTOM generates more diverse summaries than conventional product summarization.
To evaluate diversity, we follow \cite{li-etal-2016-diversity} to use Dist-2/3/4.
Since different instances of the same product have the same product information, conventional models without aspect information as input (PGen/UniLM w/o Aspect) generate the same summaries for different aspects of the same product (Top-1).
To improve the diversity of the conventional models, we also report results of conventional models without aspect information (PGen/UniLM w/o Aspect) which keep the top $K$ candidates in the beam as outputs for evaluation (Top-$K$), where $K$ is the number of instances for the product in the test set.
Experiment results are shown in Table~\ref{result_diversity}.
Our proposed CUSTOM achieves the highest performance on all diversity indicators on two datasets.
\begin{table}[h]
\centering
\small
\caption{\label{result_diversity} Diversity evaluation results. Dist-2: Distinct-2. Dist-3: Distinct-3. Dist-4: Distinct-4.}
\setlength\tabcolsep{3pt}
\begin{tabular}{lcccccc}
  \hline
 {\bf Model} & \multicolumn{3}{c} {\bf {S}{\small MART}{P}{\small HONE}} & \multicolumn{3}{c} {\bf {C}{\small OMPUTER}} \\
& Dist$-$2 & Dist$-$3 & Dist$-$4 & Dist$-$2 & Dist$-$3 & Dist$-$4 \\
  \hline 
PGen w/o Aspect (Top-1) & 0.126 & 0.187 & 0.226 & 0.034 & 0.047 & 0.056 \\
PGen w/o Aspect (Top-K) & 0.155 & 0.242 & 0.307 & 0.038 & 0.053 & 0.065 \\
PGen w/ Aspect & \bf 0.199 & \bf 0.313 & \bf 0.387 & \bf 0.107 & \bf 0.160 & \bf0.196 \\
\cdashline{1-7}[1pt/2pt]
UniLM w/o Aspect (Top-1) & 0.127 & 0.186 & 0.222 & 0.070 & 0.094 & 0.107 \\
UniLM w/o Aspect (Top-K) & 0.140 & 0.216 & 0.268 & 0.077 & 0.107 & 0.126 \\
UniLM w/ Aspect & \bf 0.232 & \bf 0.372 & \bf 0.462 & \bf 0.140 & \bf 0.205 & \bf 0.246 \\
  \hline
\end{tabular}
\end{table}

\subsection{Quality Evaluation for EXT}
Following the traditional text generation tasks, we calculate character-based {Rouge-1}, {Rouge-2} and {Rouge-L}~\cite{rouge} F1 scores to evaluate the text quality.
We use the same ROUGE-1.5.5 toolkit as in UniLM~\cite{dong2019unified}. 
We evaluate Rouge scores for overall-level and the aspect-level at the same time. Experiment results are shown in Table~\ref{results_smartphone} and Table~\ref{results_computer}.
We can conclude that, compared with the original models, both the EXT-PGen and EXT-UniLM models can obtain better Rouge scores by incorporating with the EXT framework, which verifies that EXT can boost the quality of generated aspect summaries.

\begin{table}[h]
\centering
\small
\vspace{-0.1cm}
\caption{\label{results_smartphone}Experiment results on {S}{\small MART}{P}{\small HONE}. R-1: Rouge-1. R-2: Rouge-2. R-L: Rouge-L.}
\resizebox{\textwidth}{!}{
\begin{tabular}{lcccccc}
\hline {\bf Model} & \multicolumn{1}{c} {\bf Overall} & \multicolumn{1}{c} {\bf APPEARANCE} & \multicolumn{1}{c} {\bf BATTERY} & \multicolumn{1}{c} {\bf CAMERA} & \multicolumn{1}{c} {\bf PERFORMANCE} & \multicolumn{1}{c} {\bf FEATURE} \\
& R-1/R-2/R-L & R-1/R-2/R-L & R-1/R-2/R-L & R-1/R-2/R-L & R-1/R-2/R-L & R-1/R-2/R-L\\
\hline PGen w/ Aspect & 36.3/20.4/33.4 & 37.0/20.3/33.8 & 37.7/20.0/33.9 & 39.7/23.0/37.2 & 40.7/26.9/38.3 & 28.5/13.3/25.6 \\
EXT-PGen & {\bf 37.4/21.1/34.3} & {\bf 38.7/20.9/34.8} & {\bf 39.2/22.4/35.8} & {\bf 40.1/23.0/37.4} & {\bf 42.4/27.4/39.1} & {\bf 29.3/14.5/26.9} \\
\cdashline{1-7}[1pt/2pt]
UniLM w/ Aspect & 37.3/21.5/34.2 & {\bf 38.8/22.7/35.5} & 41.6/24.5/38.2 & 38.9/21.8/35.8 & {\bf 43.0}/27.8/{\bf 39.5} & 28.7/14.7/26.3 \\
EXT-UniLM & {\bf 38.0/22.3/34.8} & 38.3/22.3/34.7 & {\bf 43.5/25.7/38.9} & {\bf 40.9/24.1/37.7} & 42.9/{\bf 27.9}/39.4 & {\bf 29.5/15.7/27.2} \\
\hline
\end{tabular}
}
\vspace{-0.5cm}
\end{table}

\begin{table}[h]
\centering
\small
\caption{\label{results_computer}Experiment results on {C}{\small OMPUTER}. R-1: Rouge-1. R-2: Rouge-2. R-L: Rouge-L.}
\setlength\tabcolsep{3pt}
\resizebox{0.8\textwidth}{!}{
\begin{tabular}{lcccc}
\hline 
{\bf Model} & \multicolumn{1}{c} {\bf Overall} & \multicolumn{1}{c} {\bf FEATURE} & \multicolumn{1}{c} {\bf PERFORMANCE} & \multicolumn{1}{c} {\bf APPEARANCE} \\
& R-1/R-2/R-L & R-1/R-2/R-L & R-1/R-2/R-L & R-1/R-2/R-L \\
\hline 
PGen w/ Aspect & 31.4/14.9/28.1 & {\bf 24.1}/9.6/21.7 & 35.5/17.7/31.7 & 34.6/17.3/30.9 \\
EXT-PGen & {\bf 32.5/16.4/29.6} & 23.9/{\bf 9.6}/{\bf 21.8} & {\bf 36.3/19.5/33.1} & {\bf 38.1/20.5/34.4} \\
\cdashline{1-5}[1pt/2pt]
UniLM w/ Aspect & 33.1/16.6/30.1 & 26.2/11.9/24.3 & 35.9/18.2/32.2 & 38.0/20.4/34.5 \\
EXT-UniLM & {\bf 33.9/17.7/31.0} & {\bf 26.8/12.6/25.0} & {\bf 36.1/19.2/32.6} & {\bf 39.8/22.3/36.4} \\
\hline
\end{tabular}
}
\end{table}

\begin{wraptable}{r}{0.6\textwidth}
	\centering
	\small
	\vspace{-1.8cm}
	\caption{\label{human}Human evaluation results.}
	\vspace{0.5cm}
	\setlength\tabcolsep{3pt}
    \begin{tabular}{lccc}
    \hline \bf Datasets & \bf Win & \bf Lose & \bf Tie \\
    \hline {S}{\small MART}{P}{\small HONE} & ${27.00\%}$ & ${11.30\%}$ & ${61.70\%}$ \\
    {C}{\small OMPUTER} & ${24.70\%}$ & ${13.00\%}$ & ${62.30\%}$ \\
    \hline
    \end{tabular}
    \vspace{-0.5cm}
\end{wraptable}

\subsection{Human Evaluation}
We conduct a human evaluation study where three participants were asked to compare summaries generated by {EXT-PGen} and {PGen} in terms of summary quality and aspect-summary consistency.
We randomly sample 100 instances for each participant from the test set for evaluation.
Table~\ref{human} shows the human evaluation results, where \textbf{Win} means the summary generated by {EXT-PGen} is better than {PGen w/ Aspect}, \textbf{Lose} means the summary generated by {EXT-PGen} is worse than {PGen w/ Aspect} and \textbf{Tie} means the summaries generated by {EXT-PGen} and {PGen w/ Aspect} obtain similar scores.
We can see that {EXT-PGen} surpasses the {PGen w/ Aspect} model at a large margin (over 10\%) on both datasets.



\subsection{Extractor Analysis}
\begin{figure*}[h]
    \centering
    \subfigure["CAMERA" aspect]{
    \begin{minipage}[b]{1\textwidth}
    \centering
    \includegraphics[width=4.6in]{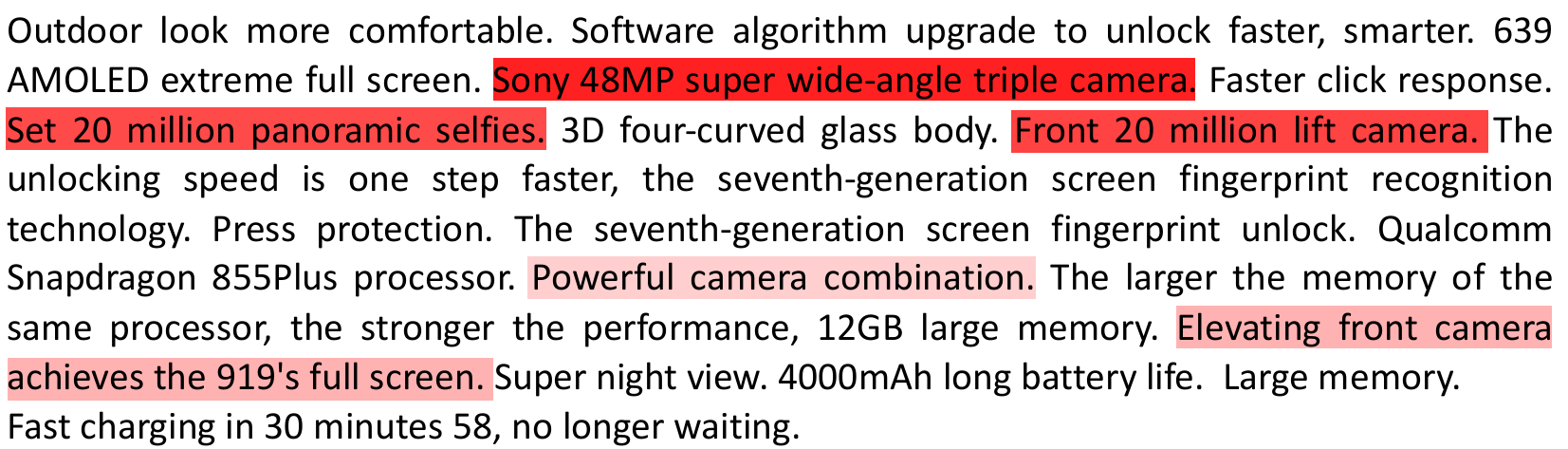}
    \end{minipage}
    }
    \subfigure["APPEARANCE " aspect]{
    \begin{minipage}[b]{1\textwidth}
    \centering
    \includegraphics[width=4.6in]{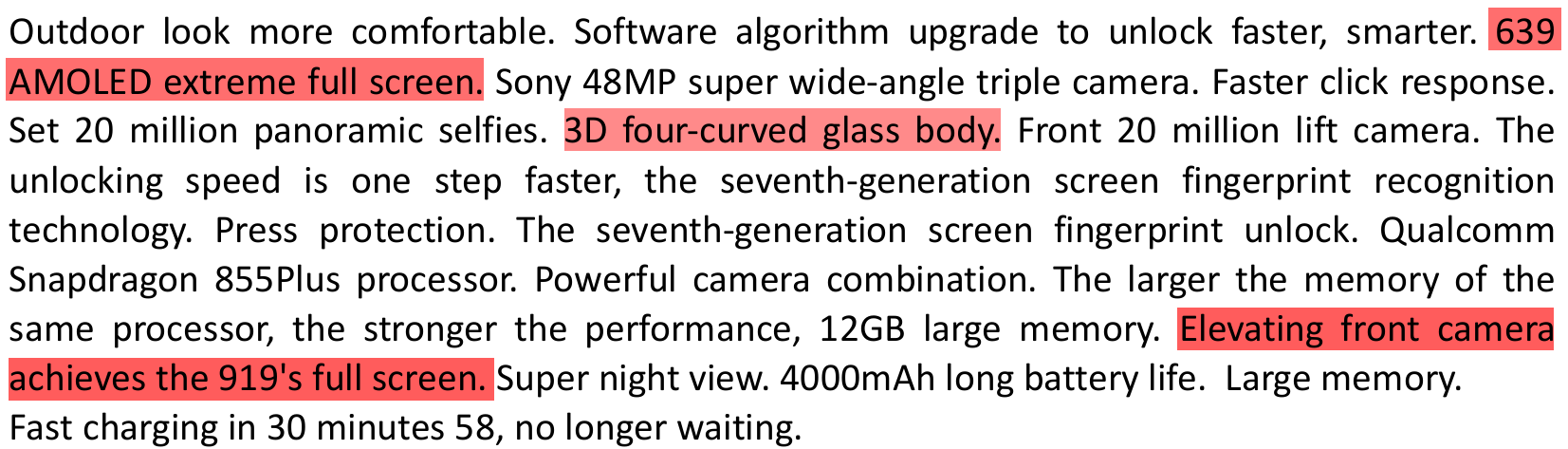}
    \end{minipage}
    }
    \caption{\label{heatmap} Heatmap of the Extractor}
\end{figure*}

We randomly select one instance from the {S}{\small MART}{P}{\small HONE} test set and plot heat maps of the extractor module.
As shown in Figure \ref{heatmap}, we find that the extractor module learned to capture sentences most relevant to the specified aspect from the product information.

\subsection{Case Study}
We perform a case study on the {S}{\small MART}{P}{\small HONE} dataset and compare the generation results of EXT-UniLM and UniLM on different aspects. Table~\ref{case_compare} shows the comparison results.
We can see that when incorporated with the EXT framework, the summary is more informative and descriptive.
For example, when describing the APPEARANCE aspect of the first instance, the EXT-UniLM generates more details, including the screen size is 6.4 inches and the screen-to-body ratio is 90.77\%. 
This shows that our EXT framework can select key information related to the current aspect from the input to improve the quality of the generated summaries.

\begin{table}
\centering
\scriptsize
\caption{\label{case_compare} Comparison of the generated aspect summaries of two models on {S}{\small MART}{P}{\small HONE} dataset. Only English translation are shown in table due to space limitation.}
\setlength\tabcolsep{3pt}
\begin{tabularx}{\textwidth}{p{0.29\textwidth}m{0.14\textwidth}<{\centering}p{0.2\textwidth}p{0.31\textwidth}}
\hline \bf Product Information & \bf Aspect & \bf UniLM & \bf EXT-UniLM \\
\hline \multirow{3}{*}{\makecell[l]{OPPO Reno 10x zoom version \\ Qualcomm Snapdragon 855 \\ 48MP ultra-clear triple camera \\ 6GB+256GB Extreme night black \\ Full Netcom \\ Full screen camera \\ Smart game phone \\ Smart protection \\ The night scene is more pure \\ Panoramic screen \\ 48 million pixels \\ Make beauty more natural \\ $\cdots$}} & \rotatebox{0}{PERFORMANCE~} & {{Equipped with Qualcomm Snapdragon 855  processor, the performance is strong, and  the operation is smooth and not stuck.}} & {{Equipped with Qualcomm Snapdragon 855  processor, it has strong performance, greatly  improves the running speed, and brings a  smoother gaming experience.}} \\
\cline{2-4}
& \rotatebox{0}{CAMERA~} & {{The rear 48 million ultra-clear three-shot,  the picture is clearer.}} &  {{The rear 48MP ultra-clear three-shot, supports 10x  hybrid optical zoom, and the photos are clearer.}} \\
\cline{2-4}
& \rotatebox{0}{APPEARANCE~} & {{Adopting a full-screen design to bring an  immersive visual experience.}} & {{This phone uses a 6.4-inch panoramic screen  design with a screen-to-body ratio of 90.77\%,  bringing you an immersive visual experience.}} \\
\hline \multirow{3}{*}{\makecell[l]{Honor 20 Youth Edition \\ AMOLED screen \\ fingerprint 4000mAh large battery \\ 20W fast charge\\ 48 million mobile phone \\ 6GB+128GB blue water emerald \\ full screen gestures \\ The night is more moving \\ Ultra-wide-angle photography \\ has a wider field of view \\ 209 slender design \\ $\cdots$}} & \rotatebox{0}{BATTERY~} & {{Built-in 4000mah high-capacity battery,  long-lasting battery life.}} & {{Built-in 4000mah high-capacity battery,  support 20w fast charge, longer battery life.}} \\
\cline{2-4}
& \rotatebox{0}{FEATURE~} & {Equipped with a full-scene acceleration engine, it  brings a smooth experience without stuttering.} & {{Equipped with a full-scene acceleration engine  and  multiple core optimizations, it brings a  smooth experience without stuttering.}} \\
\cline{2-4}
& \rotatebox{0}{APPEARANCE~} & {{With a 6.3-inch amoled pearl screen design, the screen display is clear and delicate.}} & {{With a 6.3-inch amoled pearl screen design, the  screen display is clear and vivid, bringing you an immersive viewing experience.}} \\
\hline
\end{tabularx}
\end{table}

\section{Related Work}
\subsection{Product Summarization}
Xiao et al. \cite{titlesumm} present two extractive summarization approaches using a bi-directional LSTM encoder-decoder network with attention mechanism to summarizing titles of e-commerce products.
Khatri et al. \cite{documentSummary} propose a novel Document-Context based Seq2Seq models for abstractive and extractive summarizations in e-commerce, which should be started with contextual information at the first time-step of the input to obtain better summaries. Shao et al. \cite{longDiverseGeneration} propose a Planning-based Hierarchical Variational Model to generate long and diversified expressions. They decompose long text generation into dependent sentence generation sub-tasks and capture diversity with a global planning latent variable and a sequence of local latent variables. Li et al. \cite{aspectAware} propose an Aspect-aware Multimodal Summarizer to improve the importance, non-redundancy, and readability of the product summarization.
\subsection{Conditional Text Generation}
Hu et al. \cite{controlledGeneration} propose a generative model which combines variational auto-encoders (VAEs) and attribute discriminators to produces sentences with desired attributes. Fan et al. \cite{controllableSummarization} present a neural summarization model to enable users to control the shape of the final summary in order to better suit their needs, such as selecting desired length or style. They introduce special marker tokens when training and testing.
Chen et al. \cite{kobe} propose KOBE which is the most similar task to CUSTOM.
KOBE focuses on writing expansion from short input where the generated text is likely to disrespect the truth of products, while CUSTOM concentrates on generating consistent summary from long input.

\section{Conclusion}
In this paper, we propose CUSTOM, aspe\underline{c}t-oriented prod\underline{u}ct \underline{s}ummariza\underline{t}ion for e-c\underline{om}merce, to generate diverse and controllable summaries towards different product aspects.
To support the study of CUSTOM and further this line of research, we construct two real-world Chinese commercial datasets, i.e., {S}{\small MART}{P}{\small HONE} and {C}{\small OMPUTER}.
Furthermore, we introduce EXT, an \underline{ext}raction-enhanced generation framework for CUSTOM. Experiment results on {S}{\small MART}{P}{\small HONE} and {C}{\small OMPUTER} show the effectiveness of the proposed EXT.

\section*{Acknowledgments}
We are grateful to all the anonymous reviewers. 
This work is supported by the National Key Research and Development Program of China under Grant (No.2018YFB2100802).


\bibliographystyle{splncs04}
\bibliography{custom_2}

\end{document}